# A Comprehensive Survey of Bias in LLMs: Current Landscape and Future Directions

*Rajesh Ranjan (Carnegie Mellon University, USA)*
*Shailja Gupta (Carnegie Mellon University, USA)*
*Surya Narayan Singh (BIT Sindri, India)*

**Abstract**:

Large Language Models(LLMs) have revolutionized various applications in natural language processing (NLP) by providing unprecedented text generation, translation, and comprehension capabilities. However, their widespread deployment has brought to light significant concerns regarding biases embedded within these models. This paper presents a comprehensive survey of biases in LLMs, aiming to provide an extensive review of the types, sources, impacts, and mitigation strategies related to these biases. We systematically categorize biases into several dimensions. Our survey synthesizes current research findings and discusses the implications of biases in real-world applications. Additionally, we critically assess existing bias mitigation techniques and propose future research directions to enhance fairness and equity in LLMs. This survey serves as a foundational resource for researchers, practitioners, and policymakers concerned with addressing and understanding biases in LLMs.

**Keywords**: Large Language Model (LLM), Biases, Natural Language Processing (NLP), Bias Mitigation, Artificial Intelligence and Fairness in AI.

## 1. Introduction

### 1.1 Overview of LLMs and Their Significance

Large Language Models (LLMs) have emerged as a cornerstone of contemporary natural language processing (NLP), offering transformative capabilities in text generation, comprehension, and translation. Models such as GPT-3, GPT-4, and their successors have demonstrated remarkable proficiency in generating coherent, contextually relevant text across diverse applications, including conversational agents, automated content creation, and language translation (Brown et al., 2020; Achiam et. al., 2023). These models leverage extensive datasets and advanced deep learning architectures to achieve their performance, enabling unprecedented levels of automation and efficiency in processing and generating human language (Vaswani et al., 2017). LLMs have advanced various domains, including customer service automation, educational tools, and creative industries. Their ability to understand, produce human-like text and even understand the deeper sentiments (Gupta et. al., 2024) has significantly improved user interaction experiences and expanded technological capabilities in everyday applications.

## 1.2 Motivation for Studying Biases in LLMs

Despite their impressive capabilities, LLMs are not without their challenges, notably the issue of inherent biases. Research has shown that these models can perpetuate and even exacerbate existing societal biases present in their training data (Bolukbasi et al., 2016; Caliskan et al., 2017). These biases can manifest in various forms, such as gender bias, racial bias, and contextual bias, potentially leading to unfair or discriminatory outcomes when the models are deployed in real-world scenarios (Binns et. al., 2017; Gupta et. al., 2024).

The persistence of bias in LLMs raises critical ethical and operational concerns. For instance, biased outputs from these models can adversely affect marginalized groups, contribute to misinformation, and undermine user trust (Barocas & Selbst, 2016; O'Neil et. al., 2016). Understanding and addressing these biases is crucial to ensuring that LLMs are used responsibly and equitably, fostering broader acceptance and minimizing negative impacts.

## 1.3 Objectives and Scope of the Survey

Although multiple research papers highlight the biases in LLM, the field of LLM is evolving so exponentially that a comprehensive survey of the biases incorporating a deeper fundamental understanding of biases, sources of biases in LLM, current challenges in the field, and future research direction would be a valuable handbook for researchers and practitioner. The current survey intends to fill the gap that exists today in terms of a holistic survey of biases in LLM capturing the recent advancement in the field. This survey aims to provide a comprehensive overview of biases in LLMs by systematically reviewing the existing literature and current practices related to bias detection and mitigation. Our objectives include:

1. **Categorizing Biases**: We aim to classify various types of biases observed in LLMs, such as demographic, contextual, and algorithmic biases, and analyze their sources and implications (Mehrabi et al., 2019; Zhao et al., 2019).
2. **Assessing Impact**: This survey will evaluate the impact of these biases and discuss the broader social and ethical implications (Dastin, 2018).
3. **Evaluating Mitigation Strategies**: We will review current approaches and methodologies for detecting and mitigating biases in LLMs (Zhao et al., 2019).
4. **Identifying Future Directions**: By synthesizing current research findings, we aim to identify gaps and propose future research directions to advance the field of bias mitigation in LLMs

The scope of this survey encompasses an examination of both theoretical and practical aspects of bias in LLMs to provide a holistic view of the challenges and solutions related to this critical issue.

## 2. Background

### 2.1 Evolution of LLMs: From Early Models to State-of-the-Art

The evolution of Large Language Models (LLMs) can be traced back to the early days of natural language processing (NLP), where simpler statistical models and rule-based systems dominated the field. Initial approaches relied heavily on handcrafted features and explicit programming of linguistic rules. Recurrent Neural Networks (RNNs) and Long Short-Term Memory Networks (LSTMs) introduced the ability to capture sequential dependencies and contextual information in the text (Hochreiter et. al., 1997). These models improved performance on a range of NLP tasks, including machine translation and text generation. The introduction of Transformers in 2017 (Vaswani et al. 2017) further revolutionized the field. Transformers utilize self-attention mechanisms that allow for parallel processing and improved long-range dependencies, leading to significant advancements in model scalability and performance. Models such as BERT (Bidirectional Encoder Representations from Transformers) (Devlin et al., 2018) and GPT-2 (Radford et al., 2019) demonstrated the power of pre-trained language models fine-tuned on specific tasks, setting new benchmarks across various NLP applications.

The state-of-the-art LLMs, including GPT-3 and GPT-4, represent the culmination of this evolution. These models leverage vast amounts of data and computational resources to achieve impressive capabilities in text generation, understanding, and reasoning. Current LLMs exhibit remarkable versatility in generating coherent and contextually relevant text (Brown et al., 2020). The development of these models has spurred significant research and commercial interest, highlighting their transformative impact on technology and society (OpenAI, 2023).

## 2.2 Key Concepts and Definitions

To understand biases in LLMs, it is essential to define key concepts related to bias and fairness in artificial intelligence (AI).

- **Bias**: In the context of AI and machine learning, bias refers to systematic errors or unfair tendencies in models that lead to discriminatory outcomes. Bias can arise from various sources, including biased training data, algorithmic design, and human biases (Ferrara et. al., 2023)
- **Fairness**: Fairness in AI aims to ensure that models make decisions or predictions that are equitable and do not disproportionately disadvantage any particular group. Different definitions of fairness exist, including demographic parity, equal opportunity, and individual fairness (Dastin, 2018).
- **Discrimination**: Discrimination occurs when a model's predictions or decisions unfairly disadvantage certain individuals or groups based on protected attributes such as race, gender, or age (O'Neil, 2016).

Bias and fairness are critical concerns in the deployment of AI systems. Bias in AI models can lead to harmful consequences, such as reinforcing stereotypes, perpetuating existing inequalities, and impacting marginalized communities disproportionately (Caliskan et al., 2017; Binns, 2018; Gupta et. al., 2024). The challenge of addressing bias involves identifying its sources, measuring its impact, and implementing strategies to mitigate its effects. Researchers and practitioners have developed various frameworks and metrics to assess fairness in AI systems. These include statistical measures that evaluate how predictions differ across different demographic groups and computational methods for adjusting model outputs to achieve fairness (Zhao et al., 2019; Sun et al., 2018). Ensuring fairness requires a multidisciplinary approach, incorporating insights from ethics, law, and social sciences, in addition to technical solutions.

**2.3 Types of Bias in AI-based system**

Bias can be categorized into several types based on its origin and manifestation:

- **Data Bias**: This type of bias arises from imbalances or inaccuracies in the training data. Data bias can result from underrepresentation of certain groups or overrepresentation of others, leading to skewed model predictions (Mehrabi et al., 2019). For example, if a language model is trained predominantly on text from a particular demographic, it may perform poorly on text from underrepresented groups.
- **Algorithmic Bias**: Algorithmic bias refers to biases introduced by the design of the model or the training process. This includes biases in model architecture, optimization criteria, and training procedures. For instance, algorithms that prioritize accuracy might inadvertently favor certain groups if the training data is imbalanced.
- **Systemic Bias**: Systemic bias encompasses broader societal and structural biases that influence both the data and the algorithms. This type of bias is often harder to address as it involves deep-seated social issues that are reflected in the data and perpetuated by the models (Dastin, 2018).

Understanding these types of biases is crucial for developing effective strategies to mitigate their impact and promote fairness in LLMs and other AI systems.

Large Language Models (LLMs) are susceptible to various types of biases, each with distinct implications for fairness and equity. Understanding these biases is crucial for developing strategies to mitigate their effects.

**2.3.1** Demographic bias refers to disparities in how different demographic groups are treated by LLMs. Gender bias in LLMs occurs when models exhibit differential treatment or representation of genders. For example, a model might associate certain professions with one gender more frequently than another, or generate gender-stereotyped language (Bolukbasi et al., 2016; Agiza et. al.,2024). This bias can perpetuate stereotypes and limit the perceived roles of different genders in various contexts (Zhao et al., 2019). Racial and ethnic bias involves the unfair treatment or misrepresentation of individuals based on their race or ethnicity. LLMs trained on diverse datasets might still reflect and amplify existing racial stereotypes or exhibit prejudiced behavior towards specific ethnic groups (Caliskan et al., 2017; An et. al., 2024). This can lead to harmful stereotypes and biased decision-making processes. Age bias occurs when models disproportionately favor or disadvantage individuals based on their age. For instance, LLMs might generate language that reflects age-related stereotypes or fail to account for the needs and preferences of different age groups, impacting older or younger users unfairly (Liu et. al., 2024). Bias related to socioeconomic status arises when models make assumptions or exhibit preferences based on an individual's socioeconomic background. This can affect how the model interacts with users from different economic backgrounds, potentially reinforcing socioeconomic inequalities (Singh et. al., 2024).

**2.3.2** Contextual bias refers to how LLMs produce biased outcomes based on the context in which they are applied. Certain domains, such as healthcare and finance, can introduce domain-specific biases. For example, LLMs used in healthcare might perpetuate biases present in medical literature or datasets, affecting diagnoses or treatment recommendations (Poulain et. al. 2024). In finance, models might reflect biases in credit scoring or loan approval processes (Zhou et al., 2024).LLMs trained on data from specific cultural contexts may struggle with accurately understanding or generating text for users from different cultural backgrounds. This can lead to misinterpretations or culturally insensitive outputs, impacting the inclusivity of the model (Zhang et al., 2018).

**2.4 Sources of Bias**

Bias in Large Language Models (LLMs) can stem from multiple sources throughout the development process. Understanding these sources is crucial for identifying and addressing biases effectively. The training data used to build LLMs is a major source of bias. The process of collecting and curating data for training LLMs can introduce bias if certain sources are overrepresented or underrepresented. For instance, data collected from the internet might reflect societal biases present in online content, such as gender or racial stereotypes (Caliskan et al., 2017). If the data is predominantly sourced from specific platforms or demographics, the model may not generalize well across diverse user groups. The process of annotating data—assigning labels or categories to text—can introduce bias based on the annotators' perspectives and judgments. Inconsistent or subjective annotation practices can lead to biased representations of different groups or concepts in the training data (Gautam at. al., 2024). For example, if annotators have differing views on sensitive topics, their biases may be reflected in the model's output.

The design and architecture of LLMs can also contribute to bias. Decisions made during the design phase of model development can impact how biases are represented and amplified. For example, the choice of model architecture, such as the type of attention mechanism or the depth of the network, can affect how well the model captures and mitigates biases. Some architectures may be more prone to certain types of biases due to their structural characteristics or the way they process input data. The objectives and loss functions used during training can also influence bias. Models trained with objectives that prioritize overall performance metrics, such as accuracy, may inadvertently reinforce biases present in the data. For instance, a model optimized for accuracy might perform better on majority groups while neglecting minority groups (Zhao et al., 2019).

Human factors play a significant role in introducing and perpetuating bias in LLMs. The biases of developers, researchers, and stakeholders involved in the creation and deployment of LLMs can affect the model's fairness. These biases can manifest in decisions related to data selection, feature engineering, and model evaluation. For example, if developers have implicit biases, they may unconsciously make choices that exacerbate existing inequalities or fail to recognize certain biases (Barocas et. al., 2016). Developers' cultural and societal backgrounds can also impact the design and implementation of LLMs. Different cultural perspectives may lead to varying interpretations of what constitutes fairness or bias, influencing how models are developed and evaluated. Awareness of these factors is crucial for creating more inclusive and equitable models.

| Category of Bias | Type of Bias | Description |
| --- | --- | --- |
| Application-Specific Bias | Task Bias | It occurs when the model performs differently across various tasks or domains. |
| Cognitive Bias | Confirmation Bias | The tendency to favor information that confirms existing beliefs or assumptions. |
| Data-Driven Bias | Data Bias | Arises from skewed or unrepresentative training data, leading to biased outputs. |
| Data-Driven Bias | Content Bias | Involves biases in the topics or narratives presented in the training content. |
| Data-Driven Bias | Label Bias | It occurs when the labels in the training data are biased or inconsistently applied. |
| Model Architecture Bias | Algorithmic Bias | Results from the design of the model and its learning algorithms, affecting fairness. |
| Social Bias | Gender Bias | It occurs when the model perpetuates gender stereotypes or inequities. |
| Social Bias | Racial Bias | Emerges when outputs reflect racial stereotypes or discriminatory views. |
| Social Bias | Age Bias | Reflects stereotypes associated with age, impacting representation and fairness. |
| Societal Bias | Social Bias | Reflects societal prejudices and stereotypes present in training data. |
| Societal Bias | Cultural Bias | Involves misrepresentation or underrepresentation of cultural groups. |
| Systemic Bias | Feedback Loop Bias | Arises from reinforcement of biases through user interactions and model feedback. |

Table1 : Summary of different biases in the LLM system

| Source of Bias | Description | Examples |
|---|---|---|
| Training Data | Biases originating from the data used to train the model because data may be skewed or unrepresentative. | Data from certain social media reflects societal prejudices. |
| Model Architecture | Design choices in the model that may lead to biased behavior or outputs. | Neural network configurations that amplify certain features. |
| Human Annotation | Biases are introduced during the labeling or annotation process of training data. | Inconsistent labeling of sensitive attributes like race or gender. |
| User Interactions | Feedback loops are created through user interactions that reinforce existing biases. | Users reward biased outputs, leading the model to favor them. |
| Societal Influences | Broader societal norms and stereotypes permeate the data and influence model behavior. | Reinforcement of gender roles in generated content. |
| Prompt Design | Biases arise from how prompts are structured and this can skew the model's interpretation. | Specific phrasing that leads to biased interpretations. |
| Evaluation Metrics | Metrics used to assess model performance that may not account for fairness or representation. | Using accuracy alone without considering fairness across groups. |
| Cultural Context | A lack of awareness of cultural differences can lead to misrepresentation in outputs. | Failing to recognize regional variations in language use. |
| Selection Bias | It cccurs when certain groups are overrepresented or underrepresented in the training data. | Disproportionate focus on Western cultural narratives. |

Table2: Summary of different biases in LLM system as per the source of Bias

**2.6 Impact of Bias**

The presence of bias in Large Language Models (LLMs) can have far-reaching consequences, impacting not only the performance of the models but also their societal and ethical implications. Understanding these impacts is crucial for developing strategies to mitigate bias and ensure fair and equitable AI systems. This section explores the various impacts of bias in LLMs.

**2.6.1 Social Implications:** Bias in LLMs can influence societal norms and perpetuate existing inequalities. LLMs that reflect or amplify biases against marginalized communities can exacerbate social inequalities. For example, gender or racial biases in LLMs can reinforce harmful stereotypes and perpetuate systemic discrimination. This can affect marginalized groups by reinforcing negative perceptions and limiting their opportunities in various domains such as employment, education, and healthcare (Caliskan et al., 2017). The ethical implications of bias in LLMs are significant. Models that produce biased or discriminatory outputs can contribute to ethical dilemmas, such as unfair treatment or violations of privacy. For instance, biased decision-making in sensitive areas like criminal justice or financial services can lead to unjust outcomes for individuals based on their demographic characteristics (O'Neil, 2016). Addressing these concerns requires a focus on ethical principles and the development of frameworks to ensure fairness and accountability in AI systems.

**2.6.2 Operational Implications:** Bias can also affect the operational performance and effectiveness of LLMs. Bias in training data or model design can lead to performance degradation, particularly for underrepresented or minority groups. For example, an LLM that exhibits racial bias may perform poorly in understanding or generating text related to different cultural contexts. This degradation can impact the model's overall utility and effectiveness in real-world applications (Zhao et al., 2019). Bias in LLMs can erode user trust and satisfaction. Users who encounter biased or unfair outputs may lose confidence in the technology, leading to reduced adoption and engagement. Ensuring that LLMs are fair and unbiased is crucial for maintaining user trust and satisfaction. Addressing bias through transparency, robust evaluation, and user feedback mechanisms can help improve the overall user experience and confidence in the technology.

**2.7 Bias Detection and Measurement**

Detecting and measuring bias in Large Language Models (LLMs) involves a combination of quantitative and qualitative methods. This section explores various approaches to identifying and evaluating bias.

**2.7.1** Qualitative methods provide a more nuanced understanding of bias through direct analysis of model behavior and outputs. Case studies and real-world examples help illustrate how bias manifests in specific contexts. For instance, analyzing how an LLM performs in generating content related to different demographic groups can provide insights into bias patterns. Real-world examples include instances where LLMs have been found to produce biased or discriminatory outputs in domains such as healthcare, finance, and social media. Collecting and analyzing user feedback is crucial for understanding the impact of bias from the end-user perspective. User feedback can reveal how different groups experience and perceive bias in model outputs. Techniques such as surveys, interviews, and focus groups can be employed to gather qualitative insights on user experiences and identify areas where bias may be affecting model performance or user satisfaction (Zhao et al., 2019).

**2.7.1** Quantitative methods focus on systematically measuring and evaluating bias using numerical and statistical approaches. Several metrics and benchmarks have been developed to assess bias in LLMs. Common metrics include disparity metrics (e.g., false positive/negative rates for different demographic groups), equal opportunity metrics, and fairness-aware metrics such as statistical parity and equalized odds. Benchmarks such as the Fairness Indicators and the Gender Bias Dataset provide standardized ways to evaluate bias across various applications and domains. Statistical techniques like hypothesis testing, correlation analysis, and regression analysis can reveal biases in predictions or generated text. For example, researchers might examine how different features (e.g., gender or race) influence model outcomes and identify any biased patterns. However, several composite metrics are efficient in measuring the multiple dimensions of biases. Large Language Model (LLM) Bias Index—LLMBI (Oketunji et. al., 2023), a composite scoring system created by incorporating multiple dimensions of bias including such as age, gender, and racial biases, quantify and address biases inherent in large language models (LLMs). However, LLMs still contain several implicit biases which needs to be addressed (Bai et. al., 2024). Next section focusses on recent evolution and research in the field of bias evaluation and mitigation.

## 3. Recent Bias Evaluation and Mitigation Strategies

(Lin et. al., 2024) investigate the biases present within Large Language Models (LLMs) used for bias detection, shifting the focus from solely identifying media content bias to examining biases within the models themselves. This paper analyze LLM performance in political bias prediction and text continuation tasks across diverse topics. The study also proposes debiasing methods, such as prompt engineering and fine-tuning, to mitigate bias and improve fairness in LLM outputs. D**eceiving to Enlighten** (Cheng et. al.,. 2024) is exciting concept of detecting and mitigating LLM biases. The paper emphasize the importance of equipping LLMs with mechanisms for bias recognition and self-reflection. The study demonstrates that informing LLMs their outputs don't reflect personal views, coupled with role-playing scenarios where the model self-assesses biases, leads to improved bias identification. Utilizing multi-role debate loops with an impartial referee, the proposed ranking mechanism refines outputs, outperforming existing methods in bias mitigation and contributing to ethical AI advancements. **BiasAlert** (Fan et. al., 2024), a tool designed to detect social bias in the open-text generations of Large Language Models (LLMs), integrates external human knowledge with LLM reasoning to reliably detect biases. BiasAlert outperforms state-of-the-art methods like GPT4-as-A-Judge, validating its effectiveness in evaluating and mitigating LLM biases across diverse scenarios. Social bias in code generated by Large Language Models (LLMs) is under-explored issue. A novel bias testing framework tailored for code generation tasks show 20.29% to 44.93% of generated code functions exhibit bias in sensitive tasks (Huang et. al., 2023). The paper further evaluates five bias mitigation strategies, with one-shot and few-shot learning proving most effective, removing up to 90% of bias in GPT-4's code generation. **Direct Preference Optimization (DPO)** aims to mitigate gender, racial, and religious biases in LLM-generated English text by developing a loss function that favors less biased over biased completions. The framework shows significant bias reduction and outperforms baseline models on bias benchmarks. **Social Contact Debiasing (SCD)** technique (Raj et. al., 2024) shows the possibility of reducing bias effectively. The paper uses **Contact Hypothesis** from social psychology to mitigate bias in Large Language Models (LLMs) by simulating social interactions through prompts, creating a dataset of 108,000 entries that measure 13 types of social bias across three models: LLaMA 2, Tulu, and NousHermes. The novel **Social Contact Debiasing (SCD)** technique involves instruction-tuning these models with unbiased responses. The results show that biases in LLaMA 2 can be reduced by up to 40% after just one epoch of tuning, demonstrating the effectiveness of social contact methods in addressing bias in LLM-generated content. PoliTune approach introduces a fine-tuning framework that uses **Parameter-Efficient Fine-Tuning (PEFT)** to systematically align LLMs, such as Llama3-70B, with targeted political ideologies by modifying only a small subset of parameters. This methodology provides insights into how LLMs can be tailored to reflect specific ideological biases without requiring extensive retraining, thus raising critical questions about ethical AI deployment in politically sensitive applications (Agiza et. al., 2024) The widespread integration of Large Language Models (LLMs) into software applications has brought attention to the biases these models may propagate, often stemming from the vast datasets sourced from the internet. A notable contribution in this space is LangBiTe, a platform designed to systematically evaluate bias in LLMs. LangBiTe allows developers to tailor testing scenarios and automatically generate and execute tests based on user-defined ethical requirements, offering a comprehensive traceability from ethical benchmarks to insights obtained from the model's outputs (Morales et. al., 2024). Recent research has explored how Large Language Models (LLMs) equipped with Chain-of-Thought (CoT) prompting can mitigate inherent biases during unscalable tasks such as arithmetic and symbolic reasoning. A benchmark study examining LLM predictions on gendered word lists revealed that CoT prompting reduced the social biases in LLMs (Kaneko et. al., 2024)

Recent research has highlighted that pre-trained language models (PLMs) experience "prompt bias" in factual knowledge extraction. The study demonstrates that all tested prompts exhibit notable biases, particularly gradient-based prompts like AutoPrompt and OptiPrompt, which significantly inflate benchmark accuracy by overfitting to imbalanced datasets. To address this, the authors propose a representation-based approach to mitigate prompt bias during inference. By estimating and removing biased representations, their method not only corrects overfitted performance but also enhances prompt retrieval capabilities, achieving up to a 10% absolute performance gain. This approach could serve as a benchmark for improving the reliability of PLM evaluations (Xu et. al., 2024)

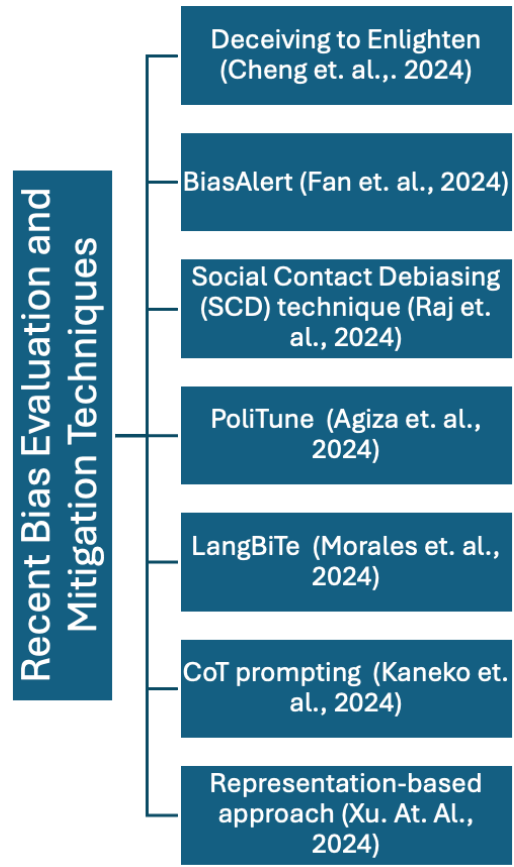

Figure1: Summary of key recent techniques for Bias evaluation and mitigation

**4. Current limitations in bias evaluation and mitigation:**

Based on the comprehensive review of biases in generative AI and LLMs, several current limitations and research gaps have emerged, which provide avenues for future exploration:

**Lack of Standardized and Intersectional Bias Evaluation Metrics:** While bias evaluation frameworks exist, they often rely on isolated metrics (e.g., gender or racial bias), leading to an incomplete view of fairness. There is a lack of comprehensive, standardized metrics that can evaluate multiple intersecting biases, such as those involving race, gender, and socio-economic status simultaneously. Additionally, these metrics are mostly domain-specific, making it difficult to apply them across different use cases and industries. Future research must develop unified, intersectional metrics for evaluating LLM biases across a range of applications, languages, and cultures.

**Bias in Underrepresented and Cross-Cultural Contexts:** Current bias research is heavily skewed towards high-resource languages and Western-centric cultural contexts, with limited focus on low-resource languages and non-Western cultures. This results in LLMs that propagate biases when applied to diverse, global user bases. Future research should focus on understanding how biases manifest in different cultural contexts, as well as in underrepresented languages, to develop fairer, more inclusive LLMs that work equitably across global contexts.

**Opaque Models and Lack of Transparency in Bias Origin:** One of the most significant gaps in bias research is the lack of transparency in the training processes and model architectures of LLMs, making it difficult to trace the origin of biases. Current models are often "black boxes," preventing researchers and developers from identifying which data points or design choices contribute to biased behavior. Addressing this requires developing more transparent AI systems and methodologies that allow for the detailed audit of data, model decisions, and training pipelines to pinpoint and mitigate bias sources.

**Scalable and Post-Training Bias Mitigation Techniques:** Existing bias mitigation techniques are often focused on pre-training solutions (e.g., data curation), but these approaches are computationally expensive and lack scalability. Additionally, post-training bias mitigation strategies, which adjust biases after the model has been deployed, are underdeveloped. This gap highlights the need for research into more scalable, cost-effective bias mitigation techniques that can be implemented both before and after training, ensuring that bias is continuously addressed throughout the model's lifecycle without requiring

## 5. Future Directions

Addressing the gaps outlined in the earlier section will improve the bias identification and mitigation capability resulting in a less biased system. Through the exhaustive analysis of the current state of the area, the following aspects are proposed as future research direction to advance the field to create more fair AI systems.

- **Comprehensive Lifecycle Bias Evaluation:** Research must expand beyond isolated bias evaluation methods and develop a holistic framework that tracks bias at every stage of the LLM lifecycle—from dataset curation, model training, and fine-tuning to real-world deployment and feedback loops. Such frameworks should integrate cross-domain metrics and provide a dynamic bias monitoring system to track changes over time, particularly as models interact with users and evolve. This will ensure continuous identification of bias across all dimensions.
- **Intersectional and Contextual Bias Mitigation:** The complexity of intersectional biases—where individuals are affected by multiple demographic attributes (e.g., gender and race)—needs to be better understood. Future research should develop sophisticated bias metrics that assess disparities across multiple demographic categories simultaneously. Additionally, more nuanced domain- and context-specific debiasing techniques are needed for high-stakes applications like healthcare and finance, where biased decisions could result in disproportionate harm.
- **Bias-Aware Training and Pre-training Techniques:** While LLMs are typically trained on vast datasets without considering bias at this stage, research should focus on developing bias-aware training mechanisms. This includes creating balanced sampling strategies and adversarial training methods during pretraining to reduce the risk of embedding societal biases into models from the outset. Bias mitigation should begin at the foundational level rather than relying solely on post-training adjustments.
- **Bias in Multimodal and Non-English Models:** As LLMs increasingly integrate multiple modalities (e.g., text, image, audio), it's essential to understand how biases propagate across modalities. Bias mitigation strategies tailored for multimodal systems, as well as those that address biases in low-resource and non-English languages, need more attention. Models trained in low-resource settings or on non-English datasets often inherit underrepresented cultural or linguistic biases, which remain a major research gap.
- **Explainability and Transparency in Bias Mitigation:** Bias mitigation methods should be accompanied by tools that make their inner workings transparent and explainable to stakeholders, including users, regulators, and researchers. Currently, many debiasing techniques function as black boxes, leaving end-users and developers unaware of how biases are mitigated. Research should prioritize building explainable models that show not only bias reduction but also how fairness adjustments influence decision-making.
- **Real-World Impact and Continuous Monitoring:** There is limited understanding of how LLM biases translate into real-world impact. Future studies should focus on conducting longitudinal, real-world assessments of LLM bias, particularly in sensitive applications like hiring, healthcare, and criminal justice. Additionally, systems should be developed that allow for real-time bias monitoring and feedback collection to adjust LLM predictions dynamically as they operate in live environments.

## 6. Conclusion:

This survey highlights the pervasive biases in Large Language Models (LLMs) and their significant implications for society. We discussed key sources of bias, including data imbalances and algorithmic disparities, and outlined effective metrics for measuring these biases. Our findings underscore the necessity of researching to identify holistic and robust bias mitigation strategies to ensure ethical AI deployment. As LLMs continue to shape critical decision-making processes, addressing these biases is essential for fostering fairness and accountability, ultimately enabling a more equitable future in AI technology.